\documentclass[runningheads]{llncs}
\usepackage{graphicx}

\usepackage[table]{xcolor}
\usepackage{listings,xcolor,amsmath}
\usepackage{pifont}
\usepackage{pdflscape}
\usepackage{multirow}
\usepackage{lscape}
\usepackage{graphicx}
\usepackage{array}
\usepackage{comment}
\usepackage{rotating,booktabs,xurl}

\usepackage{dirtree}

\definecolor{codegreen}{rgb}{0,0.6,0}
\definecolor{codegray}{rgb}{0.5,0.5,0.5}
\definecolor{codepurple}{rgb}{0.58,0,0.82}
\definecolor{backcolour}{rgb}{0.95,0.95,0.92}

\lstdefinestyle{mystyle}{
    backgroundcolor=\color{backcolour},   
    commentstyle=\color{codegreen},
    keywordstyle=\color{magenta},
    numberstyle=\tiny\color{codegray},
    stringstyle=\color{codepurple},
    basicstyle=\ttfamily\footnotesize,
    breakatwhitespace=false,         
    breaklines=true,                 
    captionpos=b,                    
    keepspaces=true,                 
    numbers=left,                    
    numbersep=5pt,                  
    showspaces=false,                
    showstringspaces=false,
    showtabs=false,                  
    tabsize=2
}

\usepackage[hidelinks]{hyperref}

\begin{document}

\title{\textsc{AutoRDF2GML}: Facilitating RDF Integration in Graph 
Machine Learning
}
\titlerunning{\textsc{AutoRDF2GML}: Facilitating RDF Integration in Graph Machine Learning}
 \author{Michael Färber\inst{1}\textsuperscript{\orcid{0000-0001-5458-8645}}
 \and David Lamprecht\inst{2}\textsuperscript{\orcid{0000-0002-9098-5389}}
 \and Yuni Susanti\inst{3}\textsuperscript{\orcid{0009-0001-1314-0286}}}

\hyphenation{Sem-Open-Alex}
\hyphenation{data-sets}

\newcommand{\orcid}[1]{\href{https://orcid.org/#1}{\includegraphics[width=10pt]{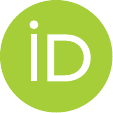}}}

\authorrunning{M. Färber et al.} %
\institute{ScaDS.AI \& TU Dresden, Dresden, Germany\\
\email{michael.faerber@tu-dresden.de}
\and
metaphacts GmbH, Walldorf, Germany\\\email{dl@metaphacts.com}
\and
Fujitsu Ltd., Japan\\\email{yuni.susanti@fujitsu.com}
}

\maketitle              %
\begin{abstract}
\label{abs}
In this paper, we introduce \textsc{AutoRDF2GML}, a 
framework designed to 
convert RDF data into data representations tailored for graph machine learning 
tasks. \textsc{AutoRDF2GML} enables, for the first time, the creation of both content-based features---i.e., features based on RDF datatype properties---and topology-based features---i.e., features based on RDF object properties. Characterized by automated feature extraction, 
\textsc{AutoRDF2GML} makes it possible even for users less familiar with RDF and SPARQL to generate data representations ready for 
graph machine learning tasks, 
such as link prediction, node classification, and graph classification.
Furthermore, 
we present four new benchmark datasets for graph machine learning, 
created from large RDF knowledge graphs using our framework. 
These datasets serve as valuable resources for evaluating graph machine learning approaches, such as 
graph neural networks. %
Overall, our framework effectively bridges the gap between the Graph Machine Learning and Semantic Web communities, paving the way for RDF-based machine learning applications.\\
\resizebox{1.02\linewidth}{!}{%
\begin{tabular}{ll}
& \\
     \textbf{Code \& Framework:}& \url{https://github.com/davidlamprecht/AutoRDF2GML/}\\
      & \href{https://opensource.org/license/mit/}{MIT License} \\

    \textbf{GML Dataset \textit{LPWC}:}& \url{https://doi.org/10.5281/zenodo.10299366} \\ &\href{https://creativecommons.org/licenses/by-sa/4.0/}{CC BY-SA 4.0 License} \\

 \textbf{GML Dataset \textit{SOA-SW}:}&  \url{https://doi.org/10.5281/zenodo.10299429} \\&\href{https://creativecommons.org/publicdomain/zero/1.0/}{Creative Commons Zero (CC0)} \\

  \textbf{GML Dataset \textit{AIFB}:}& \url{https://doi.org/10.5281/zenodo.10989595} \\&\href{https://creativecommons.org/licenses/by-sa/4.0/}{CC BY 4.0 License} \\

   \textbf{GML Dataset \textit{LinkedMDB}:}&  \url{https://doi.org/10.5281/zenodo.10989683} \\&\href{https://creativecommons.org/licenses/by-sa/4.0/}{CC BY 4.0 License} \\
\end{tabular}
}
\vspace{-0.1cm}
\end{abstract}
\vspace{-1cm}
\section{Introduction}
\label{sec:introduction}

Knowledge representation based on RDF is designed to be interpretable by both humans and machines. Integrating RDF with graph machine learning, such as in Graph Neural Network (GNN) approaches, 
however, presents significant challenges, as RDF differs remarkably from the data representations used in machine learning. 
The primary challenge lies in modeling entity relationships and attributes as 
feature vectors, diverging from RDF with its explicit knowledge representation. 
Additionally, the inherent heterogeneity 
(variety of entity and relation types) %
and sparsity of RDF data 
(few relations per entity) %
potentially affect the consistency and robustness of the learning process~\cite{world2014rdf,world2014rdfschema}.

Existing \textit{frameworks} for preparing RDF data for graph machine learning (GML) tasks typically lack the capability to transform RDF data into a propositionalized format, such as a feature matrix format. Instead, they convert RDF data into a standard feature matrix without considering the graph structure~\cite{bakhshandegan2021literal2feature}. Thus, they currently ignore both the different entity types and the object properties of RDF instances, which are crucial parts of RDF data.

Furthermore, current \textit{benchmarks} in graph machine learning, such as those provided by PyTorch Geometric, 
differ in the 
provisioning 
of node features, i.e., the modeling of nodes. 
Typically, we can categorize the available node features for datasets for graph machine learning into the following types:
(1)~content-based natural language descriptions (NLD), 
(2)~other content-based literals (e.g., numeric, categorical, or boolean values), 
and (3)~topology-based features that encapsulate the 
graph structure 
\cite{fey2019fast,lv2021we,hu2020open}.
While existing benchmarks cover both homogeneous and heterogeneous graphs, they focus on different aspects.
For homogeneous graphs, they typically prioritize the content-based features, i.e., the node features derived from natural language descriptions (NLD) of node attributes such as their labels and descriptions, while benchmarks for heterogeneous graphs typically prioritize the diversity in the graph structures or topology.
As a consequence, there is a significant gap in these benchmarks regarding the consideration of 
different kinds of semantics 
and a systematic analysis of their impact on graph machine learning models. 
This issue becomes evident when evaluating GNN-based models, as they 
frequently compute topology-based features for benchmarks that do not provide node features for all node types on-the-fly. 
Thus, analyzing whether a superior performance of a GNN-based model stems from its advanced architecture, or merely from the topology-based node features (which is then feature engineering), presents a significant challenge~\cite{li2023long}.

In this paper, we present \textsc{AutoRDF2GML}, a framework 
to effortlessly transform any given RDF data into ready-to-use heterogeneous graph datasets for graph machine learning. The generated datasets contain numeric vector features represented in feature matrices as the node features, derived from content-based (i.e., RDF datatype properties) and topology-based (i.e., RDF object properties) information of the RDF data. A notable advantage of the framework is its ability to automatically select and transform content-based features from the RDF data. 
Our framework allows users who are less familiar in RDF and SPARQL, such as those in the GNN field, to easily leverage RDF data for their research and applications. 
\textsc{AutoRDF2GML} can be installed via \texttt{pip install autordf2gml} and 
is easily set-up with a single-file configuration design: users are only required to define the RDF classes and properties, eliminating the need for specifications of complex SPARQL queries. 
Therefore, it effectively serves as a bridge between the Graph Machine Learning and Semantic Web communities, facilitating an access to a vast amount of Linked Open Data for GML purposes.

Overall, in this paper, we make the following contributions: 

\begin{enumerate}

    \item We introduce \textsc{AutoRDF2GML}, a framework to semi-automatically transform RDF data into ready-to-use heterogeneous graph datasets for graph machine learning. 
    
    \item With the proposed \textsc{AutoRDF2GML} framework, we transform four exemplary publicly available, large RDF knowledge graphs into heterogeneous graph benchmark datasets for GML (see links on page 1).

\end{enumerate}

The paper is structured as follows: In Sec.~\ref{sec:related-work}, we review related work on RDF data 
propositionalization 
and heterogeneous graph benchmarks. Sec.~\ref{sec:autordf2gml} introduces 
our framework for transforming RDF data 
for graph machine learning. 
In Sec.~\ref{sec:benchmarks}, we propose our 
RDF-based benchmarks. %
We show the potenial usage of our framework and benchmarks in Sec.~\ref{sec:usage} 
and conclude 
in Sec.~\ref{sec:conclusion}.

\section{Related Work}
\label{sec:related-work}

In this section, we first address the processing of RDF data for use in graph machine learning applications, such as graph neural networks, a process known as \textit{propositionalization}. Subsequently, we outline heterogeneous graph benchmarks.

\subsection{Propositionalization of RDF Data}
\label{sec:prepropositionalization}

Propositionalization of RDF data refers to the task of transforming raw RDF data into the format required by a given learning algorithm, such as a graph neural network~\cite{lavravc2020propositionalization}. Most data mining algorithms require a feature vector representation of the data as input, thus each instance is represented as a feature vector $(\textit{f}_1, \textit{f}_2, \dots, \textit{f}_n)$, where the features can be binary, numerical, or nominal values~\cite{ristoski2014comparison,ristoski2016semantic}.
Several approaches to generate 
such 
features from RDF data 
have been proposed. A comparison of the prominent techniques for the propositionalization of RDF data is summarized in Table~\ref{tab:PrepropositionalizationRDF}, and outlined below.

\begin{table}[tbp]
\caption{Overview prepropositionalization of RDF data.}
\begin{tabular}{llcc@{\hspace{18pt}}cc}
\toprule   
\multirow{2}{*}{} & \multirow{2}{*}{\textbf{\begin{tabular}[c]{@{}c@{}}Degree of \\ automation\end{tabular}}} & \multicolumn{2}{c}{\textbf{Feature generation}}                                                                              & \multicolumn{2}{c}{\textbf{Output}}                                                                                    \\
                  &                                                                                           & \begin{tabular}[c]{@{}c@{}}Data type \\ properties\end{tabular} & \begin{tabular}[c]{@{}c@{}}Object\\ properties\end{tabular} & \begin{tabular}[c]{@{}c@{}}Feature\\ vectors\end{tabular} & \begin{tabular}[c]{@{}c@{}}Graph structure\\ encoding\end{tabular} \\ \midrule 
Cheng et al. \cite{cheng2011automated} & non-automatic    & \ding{55}     & \ding{51}   & \ding{51}   & \ding{55}              \\
LiDDM \cite{narasimha2011liddm} & non-automatic     & \ding{51}          & \ding{55}     & \ding{51}     & \ding{55}    \\
RapidMiner \cite{khan2010two} & non-automatic   & \ding{51}        & \ding{55}      & \ding{51}     & \ding{55}     \\
FeGeLOD \cite{paulheim2012unsupervised}  & automatic    & \ding{51}        & \ding{55}    & \ding{51}   & \ding{55}    \\
Literal2Feature \cite{bakhshandegan2021literal2feature}  & automatic  & \ding{51}   & \ding{55}   & \ding{51}    & \ding{55}         \\
 \midrule
\textsc{AutoRDF2GML} & semi-automatic & \ding{51} & \ding{51} & \ding{51} & \ding{51} \\ \bottomrule
\end{tabular}
\label{tab:PrepropositionalizationRDF}
\end{table}

Cheng et al. \cite{cheng2011automated} present 
an approach for extracting features from RDF data based on user-specified feature types and SPARQL. %
The evaluation results suggest that utilizing semantic features (e.g., the taxonomy) 
improves the performance of the models in comparison to utilizing solely standard features, such as the attributes. However, unlike \textsc{AutoRDF2GML}, no content-based information (i.e., RDF datatype properties such as descriptions) is used for feature construction and the user is required to define the SPARQL queries manually.

LiDDM \cite{narasimha2011liddm} is a framework for data mining on the Semantic Web, and the data is typically retrieved via SPARQL to extract the features. %
LiDDM supports the integration of data from various Linked Open Data resources alongside a range of pre-processing techniques, including data filtering and data segmentation. However, these operations must be performed manually by the user.

RapidMiner's semweb plugin \cite{khan2010two} follows a similar approach and transforms RDF data into feature vectors, enabling its use within RapidMiner.
However, unlike \textsc{AutoRDF2GML}, the user still needs to define SPARQL queries to obtain the desired data. 

FeGeLOD \cite{paulheim2012unsupervised} and its successor, the RapidMiner Linked Open Data Extension \cite{ristoski2015mining}, are techniques for automatically enriching data with features derived from multiple Linked Open Data sources without the need to specify SPARQL queries. 
However, in contrast to \textsc{AutoRDF2GML}, their main purpose is to \textit{enrich} an existing dataset with relevant features instead of transforming RDF data into a graph dataset for graph machine learning.

Literal2Feature \cite{bakhshandegan2021literal2feature} is a framework to automatically transform RDF data into a standard feature matrix by 
traversing the RDF graph. 
It starts with a set of entities of interest and automatically retrieves literals to a pre-configured walk length to build the feature matrix. For generating the feature vectors, only the literals are used.
Literal2Feature 
is mainly used 
to obtain Spark DataFrames as input for conventional machine learning models \cite{lehmann2017distributed,draschner2021distrdf2ml}. 
In contrast to \textsc{AutoRDF2GML},
no graph structure 
is used for feature generation. %

In summary, there is currently no approach available for transforming RDF data into a propositionalized format, i.e., feature matrix, that considers both RDF data type and object properties. Our proposed framework \textsc{AutoRDF2GML} allows the representation of nodes and edges with their corresponding features as vectors. 
In addition, as with the other automatic approaches, the feature selection and transformation is performed automatically without requiring the user to define any SPARQL queries. 
The user only needs to define the key aspect of the desired GML datasets in a configuration file (e.g., the node and edge types), 
 and the rest of the processes are handled automatically.

\paragraph{Knowledge Graph Embeddings.} 
In addition to classical propositionalization methods, knowledge graph embeddings offer an approach to convert entities into dense vector representations. For instance, RDF2Vec \cite{ristoski2016rdf2vec} transforms RDF graphs into graph random walks and Weisfeiler-Lehman graph kernels, and further applies CBOW and Skip-gram models to learn latent entity representations based on the knowledge graph topology. Other graph embedding techniques includes \textit{TransE }\cite{bordes2013translating}, a translation distance model, \textit{DistMult} \cite{yang2014embedding} and \textit{ComplEx} \cite{trouillon2016complex}, semantic matching models, and \textit{RotatE} \cite{sun2019rotate}, a rotation model in the complex embedding space. 
However, these techniques 
neglect 
literals in their learning process, or only consider the literals without the topological information \cite{bakhshandegan2021literal2feature,gesese2021survey}. 

To sum up, existing frameworks to convert RDF data for graph machine learning lack an efficient method to transform the RDF data into a propositional format considering both the RDF data types and object properties of instances. \textsc{AutoRDF2GML} addresses this by facilitating the integration and utilization of both content-based (RDF data types) and topology-based (RDF object properties) embeddings, described in detail in Sec. \ref{sec:autordf2gml}.

\begin{sidewaystable}
\centering
\caption{Overview of heterogeneous graph benchmark datasets, sorted by the variety of available semantic node features.} %
\label{tab:comparisonHeteroGraphBenchmarks}
\begin{tabular}{|l|rrr|ccc|rrr|l|l|}
\hline
\textbf{\begin{tabular}[c]{@{}l@{}}Heterogeneous \\ Graph\\ Benchmark \\ Datasets\end{tabular}} &
  \multicolumn{3}{c|}{\textbf{Nodes}} &
  \multicolumn{3}{c|}{\textbf{\begin{tabular}[c]{@{}l@{}}Node Feature \\ Characteristics\end{tabular}}} &
  \multicolumn{3}{c|}{\textbf{Edges}} &
  \multicolumn{1}{c|}{\textbf{Domain}} &
  \multicolumn{1}{c|}{\textbf{\begin{tabular}[c]{@{}c@{}}Common \\ Task\end{tabular}}}\\
 &
  \textbf{\#Nodes} &
  \textbf{\begin{tabular}[c]{@{}l@{}}\#Nodes\\Types\end{tabular}} &
  \textbf{\begin{tabular}[c]{@{}l@{}}\#Node \\ Types w. \\ Features\end{tabular}} &
  \textbf{NLD} &
\textbf{\begin{tabular}[c]{@{}l@{}}Literals$\setminus$\\NLD\end{tabular}} &
   \textbf{Topology} &
  \textbf{\#Edges} &
  \textbf{\begin{tabular}[c]{@{}l@{}}\#Edge\\Types\end{tabular}} &
  \textbf{\begin{tabular}[c]{@{}l@{}}\#Edge\\Types w.\\Features\end{tabular}} &
   &
   \\ \hline
   \hline
\begin{tabular}[c]{@{}l@{}}AMiner \cite{dong2017metapath2vec} \end{tabular} &
  4,891,819 & 3 & 0 &
  \ding{55} &
  \ding{55} &
  \ding{55} &
  12,518,144 &
  2 &
  0 &
  Academia &
  Node level \\ \hline
\begin{tabular}[c]{@{}l@{}}MovieLens \cite{movielens} \end{tabular} &
  10,352 & 2 & 0 &
  \ding{55} &
  \ding{55} &
  \ding{55} &
  100,836 &
  1 &
  1 &
  Entertainment &
  Edge level \\ \hline
  \begin{tabular}[c]{@{}l@{}}LastFM \cite{fu2020magnn} \end{tabular} &
  20,612 & 3 & 0 &
  \ding{55} &
  \ding{55} &
   \ding{55} &
  128,804 &
  3 &
  0 &
  Entertainment &
  Edge level \\ \hline
HGB\_Freebase \cite{lv2021we} &
  180,098 & 8 & 0 &
  \ding{55} &
  \ding{55} &
  \ding{55}&
  1,057,688 &
  36 &
  0 &
  Multi Domain & %
  Node level \\ \hline
HGB\_LastFM \cite{lv2021we} &
  20,612 & 3 & 0 &
  \ding{55} &
  \ding{55} &
  \ding{55} &
  141,521 &
  3 &
  0 &
  Entertainment &
  Edge level \\ \hline
  Douban \cite{zhang2019inductive}&
  6,000 & 2 & 0 &
  \ding{55} &
  \ding{55} &
  \ding{55} &
  136,891 &
  1 &
  1 &
  E-commerce &
  Edge level \\ \hline
Flixster \cite{zhang2019inductive}&
  6,000 & 2 & 0 &
  \ding{55} &
  \ding{55} &
   \ding{55} &
  26,173 &
  1 &
  1 &
  E-commerce &
  Edge level \\ \hline
Yahoo-Music \cite{zhang2019inductive}&
  6,000 & 2 & 0 &
  \ding{55} &
  \ding{55} &
    \ding{55} &
  5,335 &
  1 &
  1 &
  E-commerce &
  Edge level \\ \hline
AmazonBook \cite{he2020lightgcn}&
  144,242 & 2 & 0 &
  \ding{55} &
  \ding{55} &
   \ding{55} &
  2,984,108 &
  1 &
  0 &
  E-commerce &
  Edge level \\ \hline
\begin{tabular}[c]{@{}l@{}}OGB\_MAG \cite{hu2020open}\end{tabular} &
  1,939,743 & 4 & 1 &
  \ding{51} &
  \ding{55} &
  \ding{55} &
  21,111,007 &
  4 &
  0 &
  Academia &
  Node level \\ \hline
DBLP \cite{fu2020magnn} &
  26,128 & 4 & 1 &
  \ding{51} &
  \ding{55} &
   \ding{55} &
  296,563 &
  3 &
  0 &
  Academia &
  Node level \\ \hline
IMDB \cite{fu2020magnn} &
  11,616 & 3 & 1 &
  \ding{51} &
  \ding{55} &
  \ding{55} &
  17,106 &
  2 &
  0 &
  Entertainment &
  Node level \\ \hline

HGB\_DBLP \cite{lv2021we}&
  26,128 & 4 & 3 &
  \ding{51}  &
  \ding{55
  }  &
  \ding{55} &
  239,566 &
  6 &
  0 &
  Academia &
  Node level \\ \hline
HGB\_ACM \cite{lv2021we}&
  10,942 & 4 & 3 &
  \ding{51} &
  \ding{55} &
  \ding{55} &
  547,872 &
  8 &
  0 &
  Academia &
  Node level \\ \hline
  HGB\_PubMed \cite{lv2021we}&
  63,109 & 4 & 4 &
  \ding{51} &
  \ding{55} &
  \ding{55} &
  244,986 &
  10 &
  0 &
  Academia & %
  Edge level \\ \hline
HGB\_Amazon \cite{lv2021we}&
  10,099 & 1 & 1 &
  \ding{51} &
  \ding{51} &
  \ding{55} &
  148,659 &
  2 &
  0 &
  E-commerce &
  Edge level \\ \hline
HGB\_IMDB \cite{lv2021we}&
  21,420 & 4 & 3 &
  \ding{51} &
  \ding{51} &
  \ding{55} &
  86,642 &
  6 &
  0 &
  Entertainment &
  Node level \\ \hline
H\&M \cite{hm2023}&
  1,477,522 & 2 & 2 &
  \ding{51} &
  \ding{51} &
  \ding{55} &
  31,788,324 &
  1 &
  1 &
  E-commerce &
  Edge level  \\ \hline
\cellcolor{gray!25} SOA-SW &
  170,966 & \cellcolor{gray!25} 6 &  \cellcolor{gray!25} 6 &
  \cellcolor{gray!25} \ding{51} &
  \cellcolor{gray!25} \ding{51} &
  \cellcolor{gray!25} \ding{51} &
  1,856,146 &
  7 &
  3 &
  Academia &
  Edge level \\ \hline
\cellcolor{gray!25} LPWC &
  391,247 & \cellcolor{gray!25} 4 & \cellcolor{gray!25} 4 &
  \cellcolor{gray!25}{\ding{51}} &
  \cellcolor{gray!25}{\ding{51}} &
  \cellcolor{gray!25}{\ding{51}} &
  990,303 &
  6 &
  0 &
  Academia &
  Edge level \\ \hline
\cellcolor{gray!25} AIFB &
2,260   & \cellcolor{gray!25} 3
   & \cellcolor{gray!25} 1 &
  \cellcolor{gray!25}{\ding{51}} &
  \cellcolor{gray!25}{\ding{51}} &
  \cellcolor{gray!25}{\ding{55}} &
  4,154 &
  2 &
  0 &
  Academia &
  Edge level \\ \hline
\cellcolor{gray!25} LinkedMDB &
383,290   & \cellcolor{gray!25} 8 & \cellcolor{gray!25} 8 &
  \cellcolor{gray!25}{\ding{51}} &
  \cellcolor{gray!25}{\ding{51}} &
  \cellcolor{gray!25}{\ding{55}} &
  718,675 &
  7 & %
  0 & %
  Entertainment &
  Edge level \\ \hline
\end{tabular}%
\end{sidewaystable}

\subsection{Heterogeneous Graph Benchmarks} %
\label{sec:heterogeneousGraphBenchmarks}

Several benchmarks for tasks on heterogeneous graphs (i.e., graphs with several node types) have been proposed (see Table~\ref{tab:comparisonHeteroGraphBenchmarks}). 
Our comparison includes all heterogeneous graph benchmark datasets provided by PyTorch Geometric, including heterogeneous graph benchmarks from Open Graph Benchmark (OGB) and Heterogeneous Graph Benchmark (HGB) \cite{fey2019fast,lv2021we,hu2020open}. 
For the purpose of our analysis, we categorize node features into three distinct groups: (1)~The first category encompasses \textit{natural language description (NLD)} features. These are content-based features that are derived from textual descriptions given in natural language (e.g., label or description). (2)~The second category is referred to as \textit{content-based other Literals}, denoted as \textit{Literals$\setminus$NLD}. This group includes content attributes that are not natural language descriptions, such as numeric, categorical, or boolean values. Together, both \textit{NLD} features and \textit{Literals$\setminus$NLD} constitute the broader set of \textit{content-based features (literals)}. (3)~The third category diverges from content-based attributes and is focused on the graph structure, i.e., \textit{topology}-based features.

From Table~\ref{tab:comparisonHeteroGraphBenchmarks}, it becomes apparent that existing graph benchmarks offer either content-based 
or topology-based node features across all node types, but not both. This means that most benchmark datasets focus on the heterogeneity of graph structures instead of the diversity of the node features. The benchmarks also seldom provide node features for all node types, and when they do, the generation of these features largely depends on the inherent natural language description (NLD) of the elements. 

Furthermore, 
existing benchmarks do not include separately evaluated topology-based features, which is a remarkable oversight. This is particularly relevant when new graph neural network architectures are developed, as they often calculate topology-based features on-the-fly during evaluations. Thus, it remains unclear whether the 
performance is a result of advancements in the graph neural network model itself, or due to the optimized feature engineering through the topology-based node features~\cite{li2023long}. The issues with the current benchmarks thus motivated us to construct new benchmarks datasets with our proposed \textsc{AutoRDF2GML} framework. We created \textit{SOA-SW}, \textit{LPWC}, \textit{AIFB}, and \textit{LinkedMDB} based on publicly-available RDF knowledge graphs (see Sec.~\ref{sec:benchmarks}).

\section{AutoRDF2GML} %

\label{sec:autordf2gml}

\begin{figure}[tb] %
\centering 
\includegraphics[width=1\textwidth]{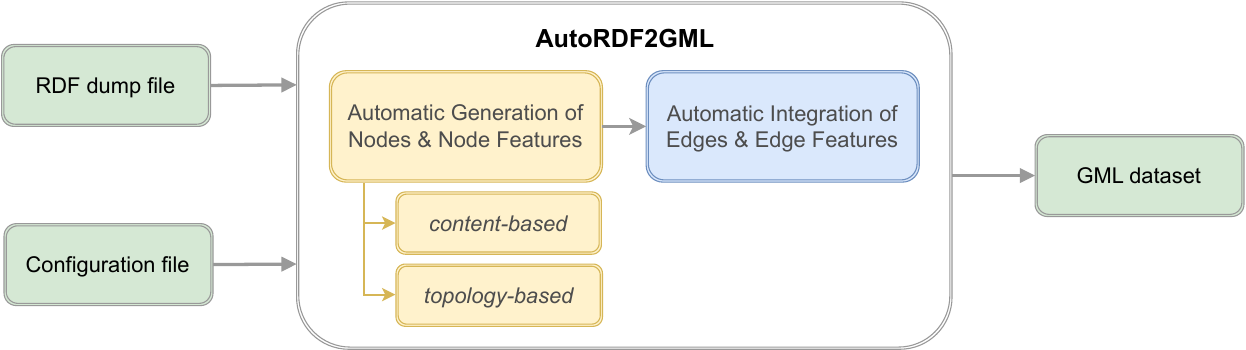}
\vspace{-0.3cm}
\caption{Overview of \textsc{AutoRDF2GML}.} 
\label{fig:OverviewAutoRDF2GML}
\end{figure}

In this section, we present our new framework \textsc{AutoRDF2GML} that seamlessly transforms RDF data into
data representations for graph machine learning tasks. %
The generated data representations contain numeric vector features represented in feature matrices as node features. 
The features can be derived from 
content-based or topology-based information in the underlying RDF data.
A notable strength of the framework is its ability to automatically select and transform the content-based features. This enable the users, even those rather unfamiliar with RDF and SPARQL, to utilize RDF data in a straightforward manner. 
The user experience is further enhanced by a  
user-centric setup: 
users are only required to define the RDF classes and 
properties for node and edge transformation, eliminating the need for complex specifications of SPARQL queries.

Figure \ref{fig:OverviewAutoRDF2GML} provides 
an overview of \textsc{AutoRDF2GML}. First, the user supplies an RDF dump file and configuration file, specifying the RDF classes and properties for feature construction.~\textsc{AutoRDF2GML} uses the \textit{rdflib} Python package, thus it supports all common RDF dump formats (e.g., Turtle, N-Triples, JSON-LD). Next, nodes are extracted from the RDF data, and their features are automatically generated based on either content-based or topology-based semantic information. Edges between the nodes are then automatically formed, completing the graph structure integration. The output of \textsc{AutoRDF2GML} is a ready-to-use heterogeneous graph machine learning dataset compatible with graph machine learning frameworks such as PyTorch Geometric~\cite{fey2019fast} and DGL~\cite{wang2019deep}.

In the following, we outline the two main steps of \textsc{AutoRDF2GML}: \textbf{(1)}~Automatic Generation of Nodes and Node Features in Sec.~\ref{autonn}, and \textbf{(2)} Automatic Integration of Edges and Edge Features in Sec.~\ref{sec:autoee}. %

\subsection{Automatic Generation of Nodes and Node Features}
\label{autonn}

In RDF data, entities belong to specific classes and are uniquely identified by URIs \cite{paulheim2012unsupervised}. Given the relevant classes specified in the configuration file, all corresponding entities are extracted to represent the nodes in the resulting graph dataset. This step is necessary for isolating the relevant classes for a specific use-case (e.g., recommendation). 
Subsequently, \textsc{AutoRDF2GML} provides two approaches to compute the node features: (1) content-based node features, and (2) topology-based node features, outlined in the following.

\subsubsection{3.1.1 Content-based Node Features.~}
After identifying the relevant entities and their corresponding URIs, \textsc{AutoRDF2GML} can generate features using RDF datatype properties. RDF datatype properties link entities to specific types of data, known as literals. These literals hold valuable information about the entity and can serve as important input features for machine learning models. 
Within its architecture, \textsc{AutoRDF2GML} includes an automatic module tailored to the construction of numeric node features based on available RDF datatype properties.
The automatic transformation of RDF datatype properties and their associated literal values into usable vectorized features includes the automatic feature selection and transformation, as outlined in the following: 

\textbf{a) Automatic Feature Selection:~}
Automating the feature selection is necessary because RDF data typically contains a huge number of datatype properties. A manual analysis and evaluation of all datatype properties is time consuming, and especially challenging for 
data scientists from other disciplines. In addition, feature selection based on RDF datatype properties is a complex task that requires addressing the following challenges:
\begin{enumerate}
    \item \textit{Property Sparsity:} The filling degree of some datatype properties can be extremely sparse.
    \item \textit{Identicality and Uniqueness of Values:} Datatype properties can include predominantly identical values or, conversely, be characterized by completely unique entries for each entity. 
    \item \textit{Redundancy:} Different properties can sometimes reflect similar information patterns, resulting in high correlation between properties. 
\end{enumerate}

We do not consider the features with any of the above listed characteristics in feature selection because they distort the underlying data dynamics, lack the necessary variance, or pose a risk of overfitting due to redundant information \cite{paulheim2012unsupervised}. Therefore, it is necessary to pre-process the available datatype properties for feature generation and only select datatype properties that do not break into any of the mentioned characteristics. 
The discarding of properties with unique values is only applied to nominal features that are not an NLD \cite{paulheim2012unsupervised}. If the values of certain selected properties strongly correlate with each other (based on the \textit{Pearson} correlation score), one of them is discarded \cite{guyon2006introduction}.

\textbf{b) Automatic Feature Transformation:~}
After the relevant features, i.e., the relevant datatype properties, are selected, they need to be transformed into a numeric vector representation to build the node features. \textsc{AutoRDF2GML} distinguishes between 6 literal types and their associate transformation rules (see Table~\ref{tab:feature-transformation}).
Strings that are natural language descriptions (NLD) are encoded using a text encoder (e.g., a language model like BERT~\cite{devlin2018bert} or SciBERT \cite{beltagy2019scibert}), following a common practice to generate node features 
\cite{hu2020open,hu2020heterogeneous,zhu2022recommender}.
Categorical values are either one-hot encoded or label encoded depending on the number of unique values. Numeric values and years are normalized. Boolean values are label encoded with 0 and 1.
For dates, the unix timestamp is calculated and then normalized. 
In general, we adhere to the established machine learning feature transformation techniques as outlined in previous works~\cite{paulheim2012unsupervised,ristoski2016semantic}. Finally, the features are normalized into a standard range, and \textit{not-a-number} (NaN) values are filled with their mean, following~\cite{guyon2006introduction}.

\begin{table}[tbp]
\centering
\caption{Transformation methods of \textsc{AutoRDF2GML}.}
\label{tab:feature-transformation}
\begin{tabular}{ll}
\toprule
\textbf{Literal Type}         & \textbf{Encoding Method}                                                         \\ 
\midrule 
String (NLD)         & Text Encoder                                                                             \\
String (Categorical) & \begin{tabular}[t]{@{}l@{}}One-Hot-Encoder\\ Label Encoder \& Normalization\end{tabular} \\
Numerical value        & Normalization                                                                            \\
Boolean              & Label Encoder (0/1)                                                                                 \\   
Year                 & Normalization   \\
Date                 & Unix Timestamp \& Normalization                                                          \\ \bottomrule
\end{tabular}
\end{table}

\subsubsection{3.1.2 Topology-based Node Features.~}
Another approach for generating features for RDF graphs 
is to leverage the topological information of the underlying RDF data.\footnote{In case both topology-based and content-based features are needed, one can run both settings and combine the features.} This is because the structure and relationships of 
the RDF object properties (e.g., relations to other entities) provide a rich semantic information about the data.
Topology-based node features are particularly flexible because they retain the complete semantic information from the topological structure, even when using a subgraph with only a small subset of the RDF classes. 
One widely used approach to obtaining the topology-based representation are knowledge graph embedding techniques. Knowledge graph embedding models such as TransE \cite{bordes2013translating}, DistMult \cite{yang2014embedding}, ComplEx \cite{trouillon2016complex}, and RotatE \cite{sun2019rotate} have gained great popularity in recent years due to their effectiveness in representing RDF entities as encoded feature vectors \cite{bakhshandegan2021literal2feature,semopenalex,ruffinelli2019you,hubert2022knowledge}. Following that, \textsc{AutoRDF2GML} automatically computes the topology-based node feature vectors using these widely recognized knowledge graph embedding techniques.

\subsection{Automatic Integration of Edges and Edge Features}
\label{sec:autoee}

To represent a complete graph structure in our data representation for Graph Machine Learning, 
we need to construct the edges. %
The edges of the transformed graph are based on RDF object properties. RDF object properties are directed relations that link two entities \cite{world2014rdf}.
Since RDF knowledge graphs may contain several object properties sharing the same range and domain and have similar semantic meanings 
(e.g. two properties linked with \texttt{owl:equivalentProperty} \cite{li2004owl}), these properties might need to be mapped to the same edge type.
Thus, \textsc{AutoRDF2GML} enables defining a list of RDF object properties that can be mapped to the same edge type. In the following, we describe four types of RDF object property patterns and how they are 
used 
within \textsc{AutoRDF2GML}.

\textbf{(a) Binary Relations:~} In RDF, an object property is a binary relation that links two entities. 
\textsc{AutoRDF2GML} creates an edge list for each edge type based on the specified RDF object properties in the configuration file. 
Although the possible subjects (i.e., start nodes) and objects (i.e., end nodes) of a relation can be defined via \texttt{rdfs:range} and \texttt{rdfs:domain} in the ontology \cite{world2014rdfschema}, our framework does not depend on this information, as we do not perform any OWL reasoning.

\textbf{(b) N-ary Relations:~}
N-ary relations \cite{giunti2021representing,world2006nary} are used when a simple binary relationship between two entities is insufficient, 
for instance when we need to model the  certainty, strength, or relevance of the relationship, as well.

\begin{figure}[tb] %
\centering 
\includegraphics[width=0.56\textwidth]{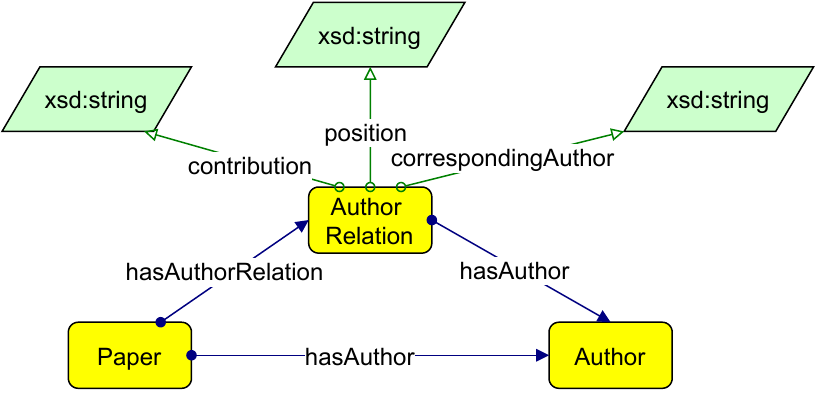}
\vspace{-0.3cm}
\caption{Example n-ary relation.} 
\label{fig:nary}
\end{figure}

The ontological pattern for modeling additional attributes that describe a relationship involves introducing a new auxiliary class. This class is linked between the subject and the object of the relationship containing additional attributes with information about the relationship \cite{giunti2021representing,world2006nary}. 
Fig. \ref{fig:nary} shows an example of n-ary relation between \textit{Paper} and \textit{Author}, where the auxiliary class \textit{AuthorRelation} contains additional information about the author's \textit{contribution} (e.g., conceptualization, supervison, or software), the author's \textit{position} (e.g., first author, middle author or last author), and if the author is the \textit{corresponding author}.

The datatype properties of the auxiliary classes that contain information about the relationships are used as edge features in the transformed heterogeneous graph dataset.
If the datatype properties do not contain numerical values, they are either one-hot encoded or label encoded.

\begin{figure}[tb] %
\centering 
\includegraphics[width=0.6\textwidth]{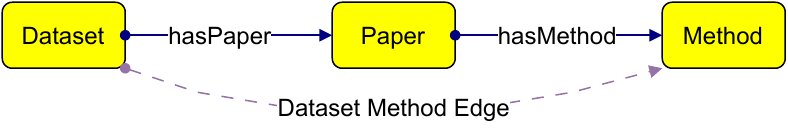} %
\vspace{-0.3cm} 
\caption{Example multi-hop relation from Linked Papers With Code.} 
\label{fig:multiHop}
\end{figure}

\textbf{(c) Multi-hop Relations:~} 
In RDF, two entities can be interconnected through a chain of object properties (see, e.g., Fig.~\ref{fig:multiHop}).
To represent these chained connections directly, edges can be formed between entities that are connected across multiple object properties.
This approach simplifies the representation and connects entities that might be several hops apart in the original RDF data \cite{world2006nary}. 
In addition, this further enables new use cases, such as link prediction across multiple properties in the underlying RDF data.
Fig.~\ref{fig:multiHop} shows an example multi-hop relation from the real-world RDF knowledge graph \textit{Linked Papers With Code} \cite{lpwc} where \textit{Dataset} and \textit{Method} are connected through the properties \textit{hasPaper} and \textit{hasMethod}. 
Given such property chain, a new edge directly connecting \textit{Dataset} and \textit{Method} is created, allowing 
the recommendation of methods for a specific dataset.

\textbf{(d) Custom Relations:~} RDF data can also contain indirect relations between classes that cannot be extracted by linear graph traveling. To extract such complex and non-linear relations and map them as explicit edges in the transformed graph, SPARQL queries can be used that explicitly define the relation.

\section{Semantic Graph Machine Learning Benchmarks}
\label{sec:benchmarks}

In the following, we present how to apply our framework to large RDF knowledge graphs, considering the semantic features of the knowledge graphs.
We provide the resulting Graph Machine Learning datasets publicly available for the 
community as benchmarks (see links on page 1). 

\subsection{SemOpenAlex-SemanticWeb (SOA-SW)} %
\label{Label2}

SemOpenAlex~\cite{semopenalex} is a vast RDF dataset containing over 26 billion RDF triples that describe 249 million publications from various academic disciplines. It features a rich schema and is interconnected with other Linked Open Data sources.

\textbf{SOA-SW Knowledge Graph Curation.~} In line with previous research on dataset curation for 
graph machine learning, we derive a subgraph of SemOpenAlex \cite{semopenalex} based on specific filter rules to create a basis for a 
graph benchmark dataset for GNN-based recommendations \cite{chennupati2021recommending,chuan2018link}.\footnotemark\footnotetext{See \url{https://github.com/davidlamprecht/semopenalex-semanticweb}} 
To ensure its validity, SOA-SW contains only SemOpenAlex entities that meet the following conditions: 
(1)~Every author included has at least one semantic web paper published and between 3 and 200 papers published in total.
(2)~From these authors, only papers with an abstract, publication year $\geq$ 2005 and citation count $\geq$ 10 are included. We exclude authors whose papers do not meet these requirements.

SOA-SW based on the SemOpenAlex version from 2023-04-24, consists of 21,978,026 RDF triples.
Table 
\ref{tab:transformed-soa-sw-nodes}
shows the number of included entities in SOA-SW. 
SOA-SW includes the comprehensive semantic information about these entities as defined in the rich SemOpenAlex ontology, covering 13 entity types, including the 
entity types \texttt{works}, \texttt{authors}, \texttt{institutions}, \texttt{sources}, \texttt{publishers} and \texttt{concepts}, as well as 87 semantic relation types \cite{semopenalex}.

\textbf{Benchmark Creation.~}For creating the benchmark, we use the SOA-SW data dump 
as input for \textsc{AutoRDF2GML}. 
We also 
add 
a custom relation to model the \textit{co-author} relations directly in the transformed graph (see GitHub).
They are not directly included in the underlying RDF data, but can be retrieved using a SPARQL query.
Modeling this relationship directly in the data allows to consider a new use case 
like collaboration recommendation. %

We construct both content-based and topology-based node features for the benchmark. For the content-based node features, the relevant data type properties are automatically selected and subsequently transformed. From all available data type properties, the following are selected: 
(a)~\texttt{Work} (8 out of 18 data type properties), (b)~\texttt{Author} (4 out of 10 data type properties), (c)~\texttt{Concept} (4 out of 11 data type properties), (d)~\texttt{Source} (10 out of 19 data type properties), (e)~\texttt{Institution} (5 out of 14 data type properties), and \texttt{Publisher} (5 out of 12 data type properties). %
Furthermore, for \textit{work-author} edges, the data type values of the property \texttt{soa:position} and for \textit{work-concept} edges, the data type values of the property \texttt{soa:score} are used as edge features.

AutoRDF2GML detects NLD features for the \textit{work} nodes, specifically the properties \texttt{dcterms:title} and \texttt{dcterms:abstract}. 
The work titles and abstracts are concatenated, and subsequently, a 128-dimensional embedding is generated for this combined data using SciBERT embeddings \cite{beltagy2019scibert}.

For the topology-based feature generation, we apply TransE~\cite{bordes2013translating}, since it yields the best results in embedding evaluation (summarized in our repository).
Using the entire SOA-SW RDF dump, AutoRDF2GML computes 128-dimensional embeddings for all nodes, utilizing all available data for the training process. 

\begin{figure}[tb] %
\centering 
\includegraphics[width=0.56\textwidth]{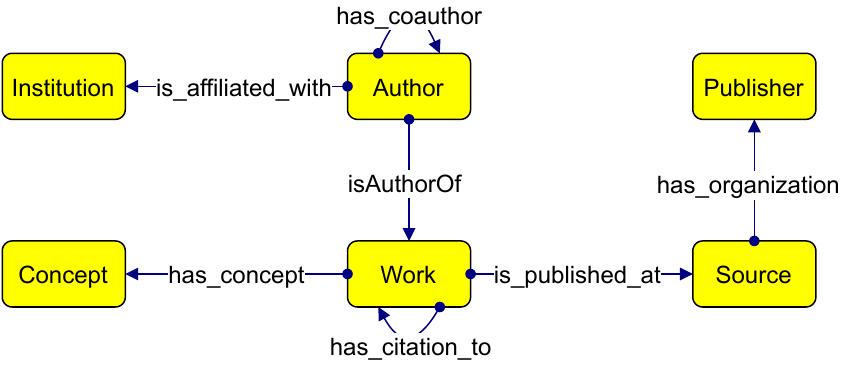} %
\vspace{-0.2cm}
\caption{Overview heterogeneous graph datasest SOA-SW.} 
\label{fig:OverviewSOA-SW}
\end{figure}

\textbf{Heterogeneous Graph Dataset SOA-SW.~}
Fig.~\ref{fig:OverviewSOA-SW} shows an overview of the schema of the created heterogeneous graph dataset based on SOA-SW KG, including in total 6 different nodes types and 7 different edge types. An overview of the number of nodes and the availability of different categories of node features for them is summarized in Table \ref{tab:transformed-soa-sw-nodes}. 
Table \ref{tab:transformed-soa-sw-edges} presents the number of edges by edge type and indicates whether they include features.
It can be seen that the transformed heterogeneous graph dataset has rich node features capturing different kinds of semantics that is available in the raw RDF data.
The semantic node features can then be used for GNN-based machine learning tasks, such as link prediction.

\begin{table}[tb]
\centering
\begin{minipage}{0.52\textwidth}
\centering
\caption{Node types, instance counts, and feature availability in \textit{SOA-SW}.}
\label{tab:transformed-soa-sw-nodes}
\resizebox{0.99\textwidth}{!}{%
\begin{tabular}{lcccc}
\hline
Node & \# Instances & Literals$\setminus$ & NLD & TransE \\
Type & & NLD & & \\
\hline
Work & 95,575 & \ding{51} & \ding{51} & \ding{51} \\
Author & 19,970 & \ding{51} & \ding{55} & \ding{51} \\
Concept & 38,050 & \ding{51} & \ding{55} & \ding{51} \\
Source & 10,739 & \ding{51} & \ding{55} & \ding{51} \\
Institution & 5,846 & \ding{51} & \ding{55} & \ding{51} \\
Publisher & 786 & \ding{51} & \ding{55} & \ding{51} \\
\hline
\end{tabular}
}
\end{minipage}%
\hfill
\begin{minipage}{0.46\textwidth}
\centering
\caption{Edge types, instance counts, and feature availability in \textit{SOA-SW}.}
\label{tab:transformed-soa-sw-edges}
\resizebox{0.99\textwidth}{!}{%
\begin{tabular}{lcc}
\hline
Edge Type & \# Instances & Features \\
\hline
work-concept & 1,320,949 & \ding{51} \\
work-source & 247,667 & \ding{55} \\
work-work & 115,271 & \ding{55} \\
author-work & 112,565 & \ding{51} \\
author-author & 38,632 & \ding{51} \\
author-institution & 19,281 & \ding{55} \\
source-publisher & 1,781 & \ding{55} \\
\hline
\end{tabular}
}
\end{minipage}
\end{table}

\begin{table}[tb]
\centering
\begin{minipage}{0.52\textwidth}
\centering
\caption{Node types, instance counts, and feature availability in \textit{LPWC}.}
\label{tab:transformed-lpwc-nodes}
\resizebox{0.99\textwidth}{!}{%
\begin{tabular}{lcccc}
\hline
Node & \# Instances & Literals$\setminus$ & NLD & TransE \\
Type & & NLD & & \\
\hline
Paper & 376,557 & \ding{51} & \ding{51} & \ding{51} \\
Dataset & 8,322 & \ding{51} & \ding{51} & \ding{51} \\
Task & 4,267 & \ding{51} & \ding{51} & \ding{51} \\
Method & 2,101 & \ding{51} & \ding{51} & \ding{51} \\
\hline
\end{tabular}
}
\end{minipage}%
\hspace{0.05\textwidth} %
\begin{minipage}{0.33\textwidth}
\centering
\caption{Edge types and instance counts in \textit{LPWC}.}
\label{tab:transformed-lpwc-edges}
\resizebox{0.99\textwidth}{!}{%
\begin{tabular}{lr}
\hline
Edge Type & \# Instances \\
\hline
paper-task & 589,784 \\
paper-method & 362,918 \\
paper-dataset & 15,520 \\
dataset-task & 14,343 \\
dataset-paper & 5,838 \\
method-paper & 1,900 \\
\hline
\end{tabular}
}
\end{minipage}
\end{table}

\subsection{Linked Papers With Code (LPWC)} %
\label{sec:use-case-lpwc}

\textit{Linked Papers With Code (LPWC)} \cite{lpwc} is an RDF knowledge graph that provides extensive information on approximately 400,000 publications in the Machine Learning field. It includes details on the tasks addressed, datasets used, methods implemented, and evaluations conducted, along with their results.
We use the \textsc{AutoRDF2GML} framework to transform 
\textit{LPWC} 
into a Graph Machine Learning dataset. 
We construct both content-based and topology-based node features. For the content-based features the following data type properties are selected:
(a)~\texttt{Paper} (3 out of 7 data type properties selected), (b)~\texttt{Method} (4 out of 7 data type properties selected), (c)~\texttt{Task} (2 out of 2 data type properties selected), (d)~\texttt{Dataset} (7 out of 10 data type properties selected).

\textsc{AutoRDF2GML} detects NLD features for all node types. %
The detected NLD features are concatenated, and a 128-dimensional embedding is calculated for the combined data with SciBERT \cite{beltagy2019scibert}. For the topology-based features, again we use TransE~\cite{bordes2013translating}, since it gives the best results in the embedding for LPWC \cite{lpwc}. 
Using the entire LPWC RDF dump, \textsc{AutoRDF2GML} computes 128-dimensional embeddings for all nodes, with all available training data.

\begin{figure}[tb] %
\centering 
\includegraphics[width=0.4\textwidth]{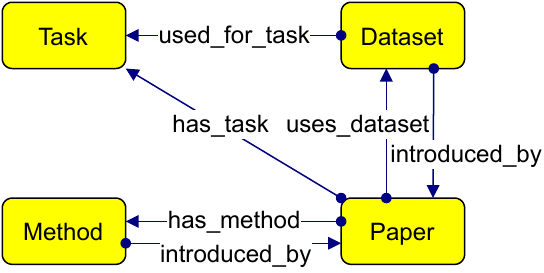} %
\vspace{-0.3cm}
\caption{Overview heterogeneous graph datasest LPWC.} 
\label{fig:OverviewLPWC}
\end{figure}

\textbf{Heterogeneous Graph Dataset LPWC.~}
Figure \ref{fig:OverviewLPWC} shows an overview of the schema of the created heterogeneous graph dataset based on LPWC, including the nodes and the edges. In total it is composed of 4 different nodes types and 6 different edge types. 
Table \ref{tab:transformed-lpwc-nodes} gives an overview of the number of nodes and the presence of different categories of node characteristics for them.
Remarkably, all node types of LPWC have node features from all three categories (Literals$\setminus$NLD, NLD and topology). 
This allows for detailed analyses of the impact of semantic node features on the performance of GNN-based machine learning tasks, such as recommendation tasks. 
An overview of the number of edges of the different edge types is shown in Table \ref{tab:transformed-lpwc-edges}.

\subsection{Further Benchmark Datasets}

We applied \textsc{AutoRDF2GML} to two other RDF knowledge graphs to show its applicability across various settings and domains. The resulting benchmarks (see page 1) were created from the RDF knowledge graphs \textbf{AIFB} \cite{aifb}, a commonly used dataset for reasoning tasks, and \textbf{LinkedMDB} \cite{linkedmdb}, the RDF version of IMDb, covering \textit{movies} and related entities such as \textit{actors} and \textit{directors}.

\section{Applications and Use Cases}
\label{sec:usage}
In this section, we outline the impact and use case examples of our framework, demonstrating its utility in both academic research and industry applications.

\textbf{Enhanced Accessibility and User-Friendliness for Semantic Data.} 
\textsc{AutoRDF2GML} enables individuals, including both established researchers and newcomers unfamiliar with RDF(S) and SPARQL, to leverage semantic web data without the need for SPARQL queries. This includes a significant number of researchers in the core Machine Learning community, including those focused on areas such as Graph Neural Networks, as well as those involved in explainable AI (XAI) and human-computer interaction (HCI). In the industry, data scientists represent a major user group for our framework, reflecting the growing demand for AI expertise.

\textbf{Increasing Use of RDF Knowledge Graphs and Linked Open Data.} 
The availability of RDF knowledge graphs, particularly in the Linked Open Data cloud, is increasing across various sectors such as e-commerce, academia, and entertainment. 
\textsc{AutoRDF2GML} supports RDF knowledge graphs without being constrained by any schema restrictions from OWL files or RDFS. 
We have established benchmarks in these domains as well, allowing for a systematic evaluation of systems such as recommender systems for products \cite{huang2004graph}, scientific papers \cite{bai2019scientific}, datasets \cite{datasetrec}, and movies \cite{jung2012attribute}. These examples not only demonstrate the availability of knowledge graphs but also an industry demand for such resources.

\textbf{Scalability and Big Data Benchmarking.} 
So far, most benchmarks have been considerably small, often consisting of only a few thousand nodes and edges (see Table 2). In contrast, \textsc{AutoRDF2GML} has been applied to 
large RDF knowledge graphs, such as LPWC with 8 million RDF triples and LinkedMDB with 6.1 million RDF triples. These benchmarks, along with others easily generated using \textsc{AutoRDF2GML}, are crucial for advancing the field and providing standardized datasets for real-world Machine Learning applications. There is a particular need for large benchmarks that include both content (node features) and structural information (topological features) to enhance AI-based systems.

\textbf{Enhancing Recommendation Systems.} 
Graph Machine Learning datasets are increasingly used in various applications, including deep learning-based search and recommender systems. Unlike systems limited to topological data and, thus, collaborative filtering approaches, the graph datasets we have developed enable more precise and higher-performing systems. Initial evaluations of recommender systems using heterogeneous graph neural networks and datasets generated with \textsc{AutoRDF2GML} have demonstrated an improvement in F1 score (see our GitHub repository). In addition, knowledge graph-based recommender systems, as summarized in~\cite{recsysSurvey10.1145/3535101}, offer several benefits. For instance, the rich semantic relationships among items in knowledge graphs helps improving item representation~\cite{Wang_2018}, and further enhances the interpretability of the recommendation results~\cite{YANG2020106194}.

\textbf{Foundation for Neurosymbolic AI.} 
\textsc{AutoRDF2GML} is well positioned to contribute to the field of neurosymbolic AI and language models. While large language models (LLMs) have been widely developed and utilized, they come with limitations such as knowledge cutoffs and significant hardware requirements. An emerging alternative involves leveraging language models, such as BERT~\cite{devlin2018bert} and T5~\cite{t5JMLR:v21:20-074} integrated with knowledge graphs \cite{kapingbaek-etal-2023-knowledge,Susanti2024,yang2023chatgpt}--an approach sometimes referred as \textit{knowledge-guided} language models. For instance, \cite{Susanti2024} introduces approach using smaller LMs combined with KGs that achieve results comparable to or even surpass those of LLM-based methods. Such approach provides capabilities in explaining the model outputs, as well, such as in recommendation systems~\cite{YANG2020106194}, by linking to KGs as explicit knowledge representations.

\section{Conclusion}
\label{sec:conclusion}

In this paper, we introduced \textsc{AutoRDF2GML}, a novel framework designed to efficiently convert RDF data into benchmarks tailored for graph-based machine learning applications, potentially bridging the gap between the Graph Machine Learning and Semantic Web communities. \textsc{AutoRDF2GML} framework is characterized by its modular design, automated feature extraction, and one-file configuration design, 
making it accessible even to users who may not be familiar with semantic technologies such as SPARQL. 
Furthermore, we demonstrated the utility of \textsc{AutoRDF2GML} by applying it to large 
RDF knowledge graphs, successfully transforming them into heterogeneous graph datasets, each enriched with unique semantic features.

In the future, we plan to enhance our framework to operate across multiple RDF knowledge graphs within the Linked Open Data cloud in parallel and to incorporate reasoning through OWL concepts.
This enhancement will include mechanisms for handling ontological relationships across different knowledge graphs, such as \texttt{equivalentClass} 
links.

\paragraph*{Resource Availability Statement:} All resources are accessible through the URLs provided on page~\ref{abs}.

\bibliographystyle{splncs04}
\bibliography{references}

\end{document}